\documentclass{article}

\usepackage{arxiv}
\usepackage{authblk} 
\usepackage{placeins}
\usepackage[utf8]{inputenc}
\usepackage[T1]{fontenc}
\usepackage{hyperref}
\usepackage{url}
\usepackage{booktabs}
\usepackage{amsfonts}
\usepackage{nicefrac}
\usepackage{microtype}
\usepackage{lipsum}
\usepackage{graphicx}
\usepackage{multicol}
\usepackage{amsmath}
\usepackage[table]{xcolor}
\graphicspath{{./images/}}

\title{Exploring Major Transitions in the Evolution of Biological Cognition With Artificial Neural Networks}

\author[1,2]{Konstantinos Voudouris}
\author[3]{Andrew B. Barron}
\author[4]{Colin Klein}
\author[5,2]{Marta Halina}
\author[6,7]{Matishalin Patel}

\affil[1]{Institute for Human-Centered AI, Helmholtz Zentrum Munich, Neuherberg, Germany (\texttt{kv301@srcf.net})}
\affil[2]{Leverhulme Centre for the Future of Intelligence, University of Cambridge, Cambridge, UK}
\affil[3]{School of Natural Sciences, Macquarie University, Sydney, Australia}
\affil[4]{School of Philosophy, The Australian National University, Canberra, Australia}
\affil[5]{Department of History and Philosophy of Science, University of Cambridge, Cambridge, UK}
\affil[6]{School of Environmental and Life Sciences, University of Hull, Hull, UK}
\affil[7]{Centre for Data Science, AI, and Modelling, University of Hull, Hull, UK}

\begin{document}
\maketitle
\begin{abstract}
Transitional accounts of evolution emphasise a few changes that shape what is evolvable, with dramatic consequences for derived lineages.  More recently it has been proposed that cognition might also have evolved via a series of major transitions that manipulate the structure of biological neural networks, fundamentally changing the flow of information. We used idealised models of information flow, artificial neural networks (ANNs), to evaluate whether changes in information flow in a network can yield a transitional change in cognitive performance.  We compared networks with feed-forward, recurrent and laminated topologies, and tested their performance learning artificial grammars that differed in complexity, controlling for network size and resources. We documented a qualitative expansion in the types of input that recurrent networks can process compared to feed-forward networks, and a related qualitative increase in performance for learning the most complex grammars. We also noted how the difficulty in training recurrent networks poses a form of transition barrier and contingent irreversibility --- other key features of evolutionary transitions. Not all changes in network topology confer a performance advantage in this task set. Laminated networks did not outperform non-laminated networks in grammar learning. Overall, our findings show how some changes in information flow can yield transitions in cognitive performance. 
\end{abstract}


\section*{Introduction}

Inspired by work on major transitions in the history of life \cite{SmithThe-major97}, there have been several recent proposals for major transitions in the evolution of cognition \cite{barron2023transitions,bennett2023brief,GinsburgThe-Evolution19}. Transitional accounts are attractive because they offer the promise of organising a bewildering variety of cognitive changes around a small set of key innovations. A major transition differs from other important evolutionary changes in several respects \cite{SmithThe-major97,CalcottThe-major11}. First and foremost, major transitions change the space of what can evolve given realistic constraints. These changes in \emph{evolvability} \cite{BrownWhat14} thus have important impacts on what is feasible for downstream lineages. Furthermore, while transitional forms are possible, they are unstable: there are fitness advantages on either side of a transition, but the gap is hard to cross and hard to undo. Hence transitions are contingently irreversible once they happen. Transitional accounts face a core challenge. While a transition changes evolvability, the evolutionary explanation for a transition cannot make reference to evolvability: any benefits are obtained by descendants, and much later than the transition itself. A transition does not occur because of the benefits it will later confer. It must be explained through other means.

Rather than focus on specific cognitive abilities (such as navigation or imagination \cite{bennett2023brief,GinsburgThe-Evolution19}), we have argued that a useful way to think about cognitive transitions is in terms of changes to the \emph{computational architecture} of a system \cite{barron2023transitions,klein2024comparing}. The computational architecture of a cognitive system is the capacities it has for operations, representation, and sequencing control flow. Changes in computational architecture can be favoured on the grounds of resource consumption when they allow the same tasks to be done more efficiently \cite{KleinExplaining22}, while enlarging the class of problems that systems of the same type can solve \cite{barron2023transitions,klein2024comparing}. 

A key piece of this argument is that changes in the structure and organisation of nervous systems, and more particularly changes in large-scale patterns of neural connectivity, would be sufficient to change computational architecture. There is comparative evidence that recurrence (the feedback of later neural processing to earlier stages) and lamination (splitting processing into multiple distinct streams) correspond to turning points in cogntive evolution \cite{barron2023transitions}.  There is also evidence that different artificial neural networks  with distinct topologies differ in the classes of formal languages that they can recognise  \cite{deletang2022neural,weiss2018practical,ackerman2020survey,giles1992learning, elman1990finding,gers2001lstm}. Although both lines of evidence are suggestive, they are also indirect. 

We used modelling to look at the effect of changes in computational architecture on cognition.  We focussed on the problem domain of artificial grammar learning, for which the computational complexity of a problem can be precisely stated. We explored a granular hierarchy of problems described by the sub-regular formal language hierarchy \cite{rogers2011aural} and the Chomsky hierarchy \cite{chomsky1956three,chomsky1959algebraic,chomsky1959certain}. Such tasks have been used elsewhere as idealized models of the evolution of language \cite{fitch2004computational,gentner2006recursive, abe2011songbirds}, but we used them here as representatives of a general hierarchy of increasingly complex problems.  

We generated artificial neural networks in a way that allowed precise control of their topology, scale, and neural resources. We focussed on the differences in performance between feed-forward networks and recurrent networks (both standard recurrent architectures and architectures with gated-recurrent units) as well as architectures that were either fully connected or have laminated partitions in their hidden layers.

\begin{figure}[t!]
\centering
\includegraphics[width=\linewidth]{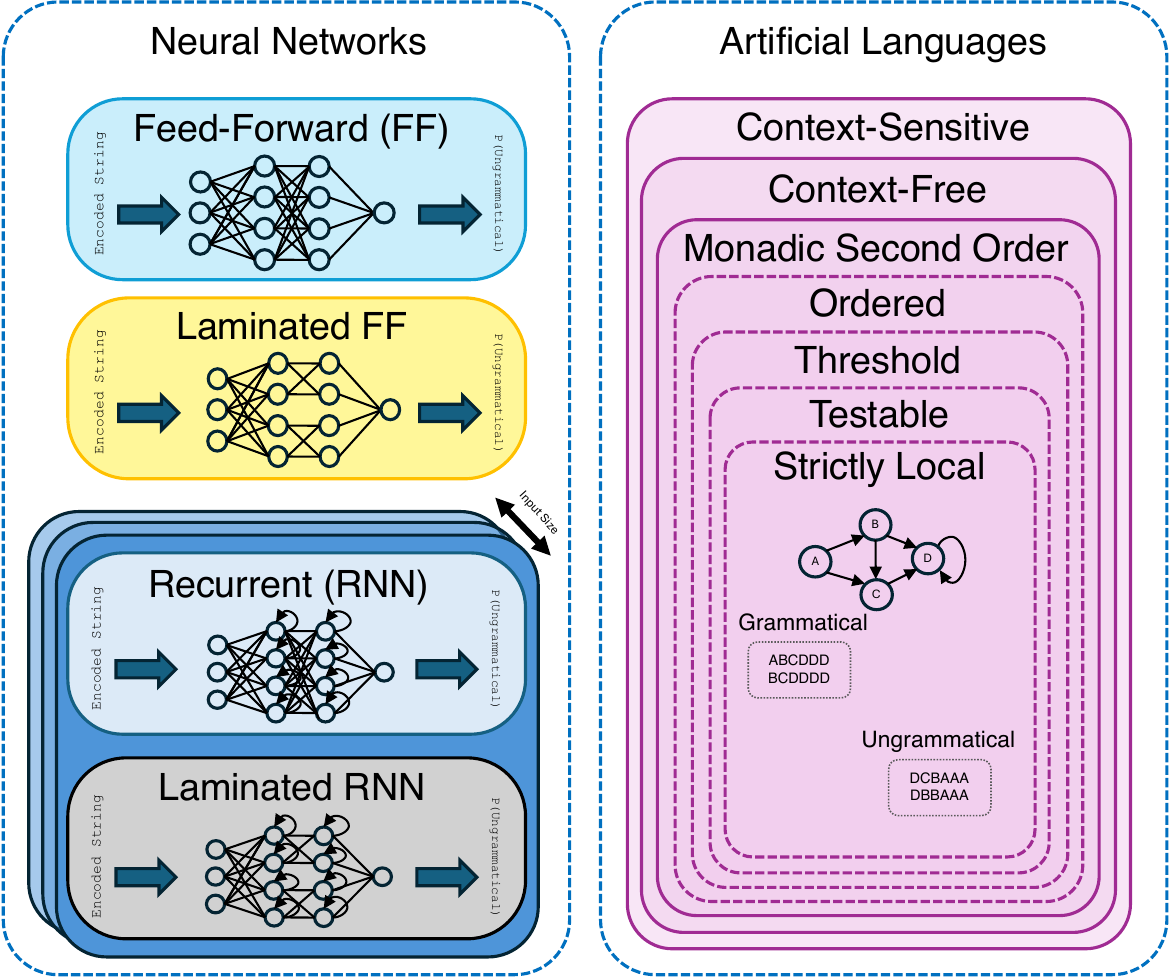}
    \caption{
    We compared the performance of ANNs with different topologies inspired by proposed major transitions in cognitive evolution on the principled problem of artificial grammar learning. Lamination was interpreted as partitioned hidden layers which are combined in the final layer. Recurrence was implemented as either a simple hidden state network or using Gated Recurrent Units (GRU) layers. The grammar hierarchy is a expanded form of the Chomsky Hierarchy where we have subdivided the regular grammars into separate classes of the sub-regular grammars to allow a finer resolution in complexity.}\label{fig:overview}
\end{figure}

These models idealise both the information flow in actual neural networks and the complexity of actual learning tasks, and allow us to ask precise questions about whether changes in functional network structure can yield performance differences that appear to have transitional properties. We trained each network to discriminate grammatical and ungrammatical strings generated from an artificial grammar of a given complexity, and varied the number of characters they could see at a given processing step (Figure \ref{fig:overview}).  By comparing the differences in performance across the hierarchy of artificial grammars between networks of different types, we were able to ask whether a structural change in cognitive architecture can yield a transition in cognitive performance.

\section*{Results}

We generated grammars for every level in the sub-regular hierarchy, as well as grammars for context-free and context-sensitive languages, and used them to generate 500 grammatical and 500 ungrammatical 12-letter strings from a six-letter alphabet. The grammars and their abbreviations are described in Table \ref{tab:grammars}. We then trained neural networks on 800 randomly selected strings, and tested them on the remaining 200. We exhaustively generated neural networks with three architecture types, feed-forward networks (FFNs), recurrent neural networks (RNNs), and gated recurrent unit neural networks (GRUs), with 32 to 512 neurons (in increments of 32), with and without lamination in the hidden layers, and up to three layers deep. GRUs are a type of recurrent architecture that improve on the training stability of RNNs \cite{cho2014properties}.

\begin{table}[t!]
\rowcolors{1}{white}{gray!15}
\centering
\caption{The grammars in the sub-regular and Chomsky hierarchies used in this study.}
\begin{tabular}{p{2.1in}p{2in}}
Grammar & Description  \\
\midrule
Strictly $k$-Local (SL) & Only some $k$-size sequences can be present. \\
Locally $k$-Testable (LT) & Some $k$-size sequences must be present. \\
Locally $t$-Threshold $k$-Testable (LTT) & Some $k$-size sequences must be present $<t$ times. \\
LTT With Order (LTTO) & Some $k$-size sequences must be present in some order. \\
$n$-Monadic Second Order (MSO) & Some subsequences must be present $\text{mod} ~n$ times. \\
Context-Free (CF) & Mirrored or repeated subsequences, or $A^nB^n$ languages with valid transitions between characters in $A$ and $B$. \\
Context-Sensitive (CS) & $A^nB^nC^n$ languages with valid transitions between characters in $A$ and $B$, and $B$ and $C$. \\
\bottomrule
\end{tabular}
\label{tab:grammars}
\end{table}

Feedforward networks classify strings by taking in the whole string at once and produce a probability that the string is grammatical or ungrammatical. Recurrent architectures (RNNs and GRUs) process strings as a sequence, taking in one character at a time and updating their hidden layer weights. Once the whole string is processed, the stored layer weights are used to produce a probability that the string is grammatical or ungrammatical. We varied the number of characters (the input size) that the recurrent architectures see in a single pass, ranging from 1 to 12 (the length of the string). The network moves through the string with a sliding window of the corresponding size.

To explore the differences between each of these variables, we ran a fixed effects beta regression analysis on the proportion of test strings that each architecture correctly classified. We included main effects for the number of neurons, the number of layers, the architecture type, whether the model is laminated or not, the grammar type, and the input size (fixed at 12 for FFNs). We also included interaction effects for architecture type and grammar type, lamination and grammar type, architecture type and lamination, and input size and grammar type. This was the best fitting model based on Akaike and Bayesian Information Criteria, and 5-fold cross-validation. Using the fitted regression model, we could estimate the predicted percentage of test strings correctly classified (estimated marginal means) \textit{after} controlling for confounding variables, such as number of neurons, number of layers, and grammar type.

\begin{figure}
\centering
\includegraphics[width=\textwidth]{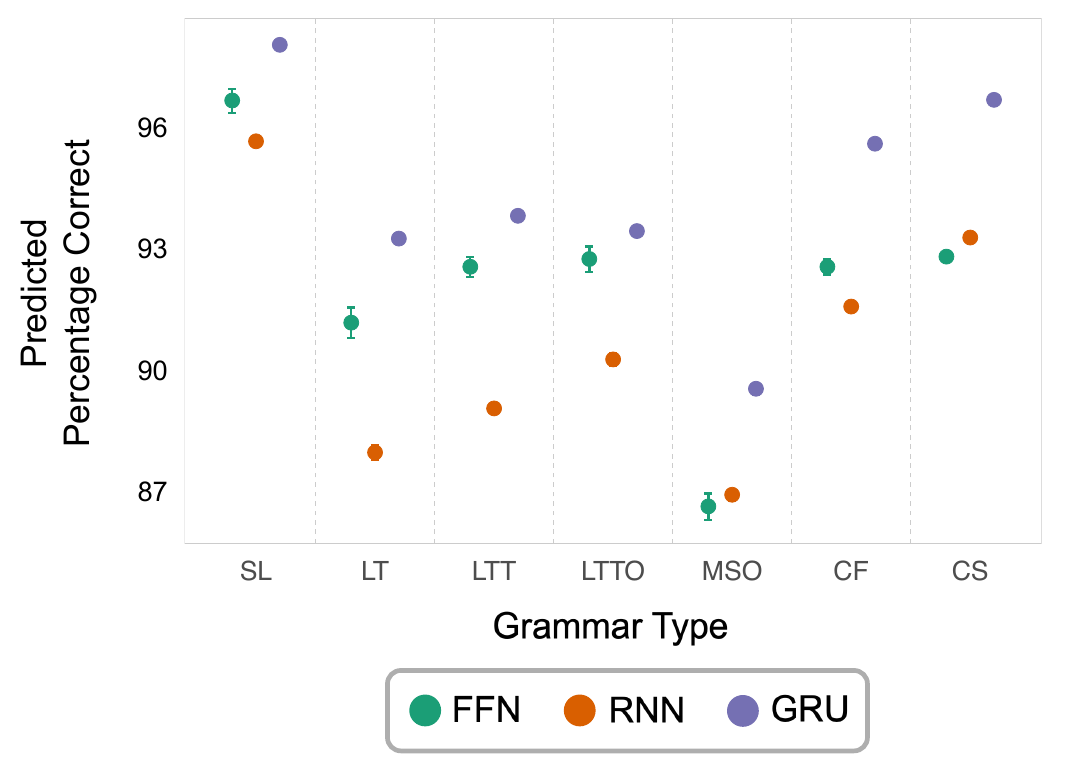}
    \caption{\textbf{Gated Recurrent Unit Neural Networks were significantly better than feed-forward architectures on most problems.} The predicted percentage of test strings that each architecture type correctly classified, with 95\% confidence intervals, estimated using our beta regression model. Feedforward networks (FFN) were compared to two recurrent architectures, Recurrent Neural Networks (RNN) and Gated Recurrent Unit Neural Networks (GRU). They were tested on seven grammar types: strictly (SL), locally testable (LT), locally threshold-testable (LTT), locally threshold-testable with order (LTTO), monadic second order (MSO), context-free (CF) and context-sensitive (CS). }\label{fig:grammars-by-architecture}
\end{figure}

\subsection*{Recurrence Improves Performance for Harder Problems}

Figure \ref{fig:grammars-by-architecture} presents the predicted percentage of strings in the held-out test set that each architecture type was estimated to correctly classify on each grammar type, holding all other covariates constant. The grammars are presented in ascending order, from left to right, on the sub-regular and Chomsky hierarchies. The relatively poor performance of RNNs across these tasks was likely due to the instability of training sequence models with ordinary recurrent connections, a problem which is remedied by introducing gated recurrent units in GRUs (more on this below).

\begin{table}[h]
    \rowcolors{1}{white}{gray!15}
    \centering
    \caption{The predicted percentages for the three architecture types, with differences between FFNs and the recurrent architectures shown in brackets.}
    \begin{tabular}{llll}
    Grammar & FFN & RNN & GRU \\
    \midrule
    SL & 96.49 & 95.52 (--0.97) & 97.81 (+1.32)\\
    LT & 91.20 & 88.12 (--3.09) & 93.20 (+2.00) \\
    LTT & 92.53 & 89.16 (--3.37) & 93.74 (+1.21) \\
    LTTO & 92.72 & 90.33 (--2.38) & 93.38 (+0.66) \\
    MSO & 86.83 & 87.11 (--0.28) & 89.63 (+2.80) \\
    CF & 92.53 & 91.59 (--0.95) & 95.46 (+2.93) \\
    CS & 92.77 & 93.23 (--0.45) & 96.50 (+3.73) \\
    \bottomrule
    \end{tabular}
    \label{tab:recurrence-results}
\end{table}

Table \ref{tab:recurrence-results} presents the predicted percentages for FFNs, RNNs, and GRUs. GRUs were significantly better than both FFNs and RNNs on all grammar types ($p<0.0001$). RNNs were significantly worse than FFNs on all grammar types ($p<0.0001$), with the exception of the Monadic Second Order languages ($p=0.2787$). The difference between the predicted percentages for FFNs and GRUs was larger for more complex grammars. We found a similar pattern when comparing RNNs and FFNs. While RNNs were consistently outperformed by FFNs, the difference between them decreased from the Monadic Second Order languages onwards.

In general, recurrence with the addition of gated units to improve training stability conferred a significant advantage across all grammar types. Moreover, the advantage of recurrent architectures was greater for more complex problems, exemplified by the Monadic Second-Order, Context-Free, and Context-Sensitive languages. This improvement was seen for both ordinary Recurrent Neural Networks and Gated Recurrent Unit Neural Networks.

\subsection*{Lamination is a Disadvantage on Harder Problems}

\begin{figure}[h]
\centering
\includegraphics[width=\textwidth]{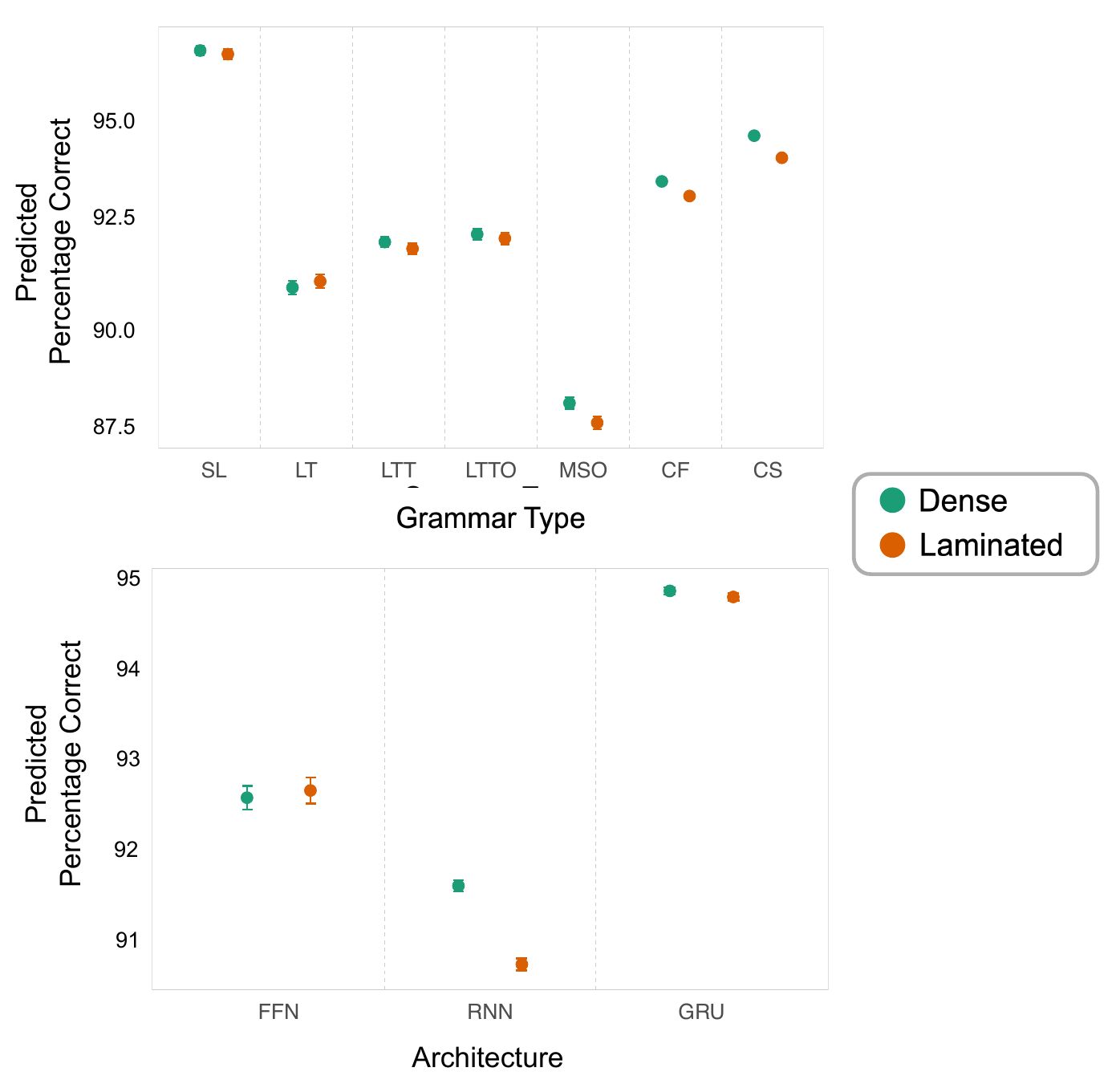}
    \caption{\textbf{Lamination did not confer a significant advantage on grammar learning problems.} The predicted percentage of test strings that dense and laminated networks correctly classified, with 95\% confidence intervals, estimated using our beta regression model. The top panel presents the effect of lamination across seven grammar types: strictly local (SL), locally testable (LT), locally threshold-testable (LTT), locally threshold-testable with order (LTTO), monadic second order (MSO), context-free (CF) and context-sensitive (CS). The bottom panel presents the effect of lamination across three architecture types: Feed-Forward Networks (FFN), Recurrent Neural Networks (RNN), and Gated Recurrent Unit Neural Networks (GRU).
    }\label{fig:laminations-by-grammar-by-architecture}
\end{figure}

Figure \ref{fig:laminations-by-grammar-by-architecture} presents the predicted percentage of strings in the held-out test set that dense versus laminated networks were estimated to correctly classify on each grammar type (top) and each architecture type (bottom), holding all other covariates constant. The difference between dense and laminated architectures was minimal across all conditions, with a slight, but significant, disadvantage conferred by introducing lamination. 

\begin{table}[h]
    \rowcolors{1}{white}{gray!15}
    \centering
    \caption{The predicted percentages for dense and laminated architectures across grammar and architecture types, with p-value significances for the differences in parentheses.}
    \begin{tabular}{llllr}
     Main Effect & Factor & Dense & Laminated & $p$-value \\
    \midrule
    Grammar & SL & 96.79 & 96.70 (--0.09) & 0.1642 \\
    & LT & 90.99 & 91.14 (+0.15) & 0.1052 \\
    & LTT & 92.10 & 91.94 (--0.16) & 0.0172 \\
    & LTTO & 92.29 & 92.19 (--0.10) & 0.1758 \\
    & MSO & 88.16 & 87.68 (--0.48) & $<$0.0001 \\
    & CF & 93.59 & 83.23 (--0.36) & $<$0.0001 \\
    & CS & 94.70 & 94.16 (--0.54) & $<$0.0001 \\
    \midrule
    Architecture & FFN & 92.56 & 92.64 (+0.08) & $<$0.0001 \\
    & RNN & 91.59 & 90.73 (--0.86) & $<$0.0001 \\
    & GRU & 94.83 & 94.76 (--0.07) & $<$0.0001 \\
    \bottomrule
    \end{tabular}
    \label{tab:lam-results}
\end{table}

Table \ref{tab:lam-results} presents the predicted percentages for dense versus laminated architecture types and the significance of the differences between them, across grammar and architecture types. As with the pattern in architecture types, there was a difference between simpler and more complex languages. Although the laminated networks were generally worse than dense networks, that difference was small and not statistically significant for the SL to LTTO levels of the grammar hierarchy. However, from the monadic second-order languages onwards, the difference increased and became significant.


The effect of lamination across architectures also differed. For Feed-Forward Networks, lamination conferred a small but statistically significant \textit{advantage}. In contrast, lamination conferred a significant \textit{disadvantage} on both Recurrent Neural Networks and Gated Recurrent Unit Neural Networks. Indeed, this disadvantage was largest for RNNs, likely due to the increased training instability from having partitioned hidden layers, worsening an already unstable training regime.

Lamination, in the sense of partitioned hidden layers in an artificial neural network, appeared to be disadvantageous for learning to differentiate grammatical and ungrammatical strings. This disadvantage was amplified for more complex grammars. There is nonetheless some evidence that this disadvantage mainly affects recurrent architectures, with Feed-Forward Networks benefitting slightly from a laminated topology.

\subsection*{Input Size Significantly Affects Performance}

\begin{figure}[h]
\centering
\includegraphics[width=\textwidth]{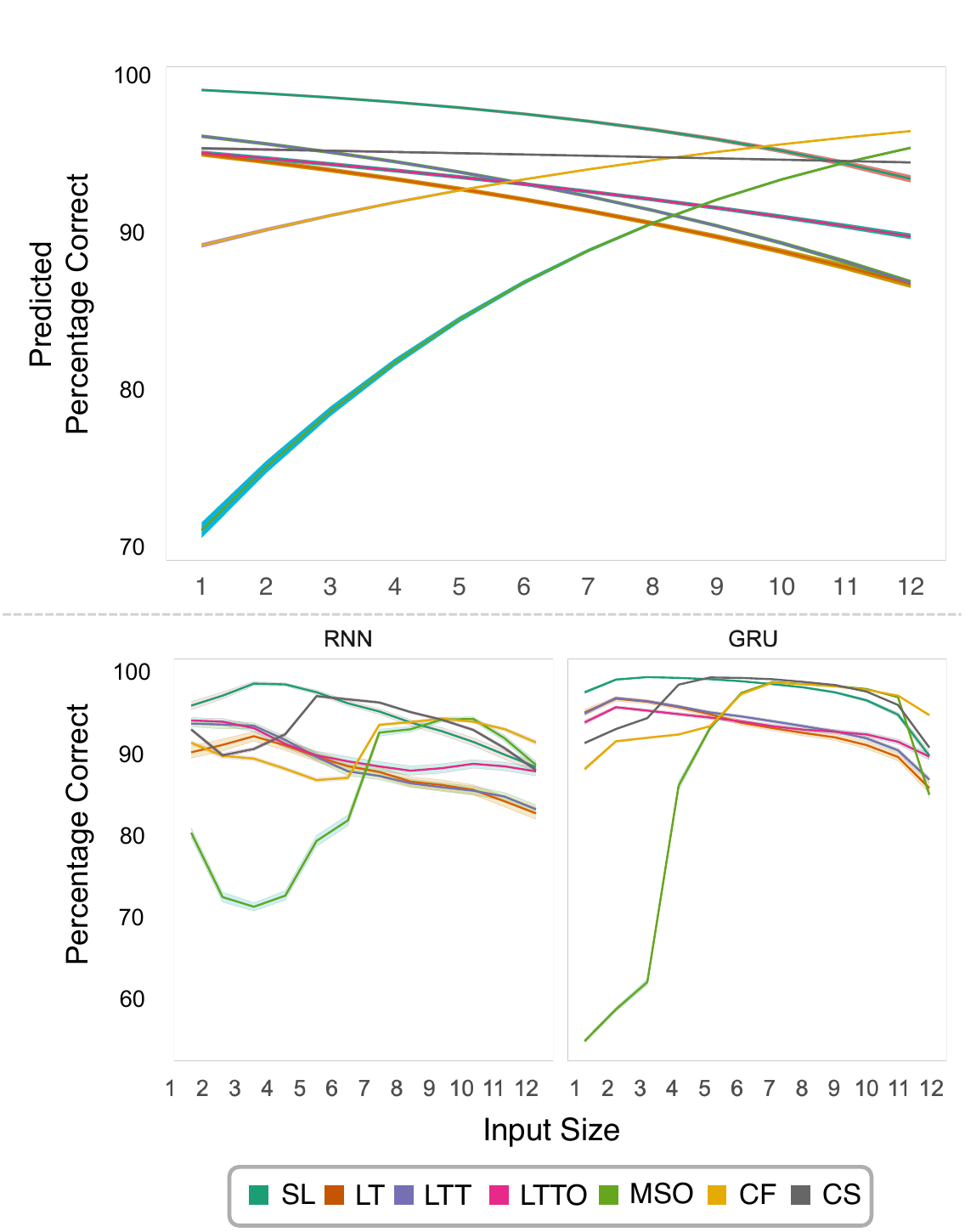}
    \caption{\textbf{Input size has a significant effect on the performance of recurrent architectures.} The percentage of strings that recurrent architectures correctly classify depending on the number of letters (input size) they receive at each weight update step. The top panel presents the predicted percentage correct based on our regression model for each grammar type: strictly local (SL), locally testable (LT), locally threshold-testable (LTT), locally threshold-testable with order (LTTO), monadic second order (MSO), context-free (CF) and context-sensitive (CS). The bottom left panel presents the raw mean percentage correct for Recurrent Neural Networks (RNN) and the bottom right panel shows the same for Gated Recurrent Unit Neural Networks (GRU). All plots are presented with 95\% confidence intervals.}\label{fig:inputsize-by-grammar-by-architecture}
\end{figure}

Since the recurrent architectures process strings as sequences, we varied the number of letters that they processed at each step through the sequence. We found that this had a significant effect on performance. Figure \ref{fig:inputsize-by-grammar-by-architecture} presents the predicted percentage, as well as the raw unadjusted mean percentage, of strings in the held-out test set for each input size for each grammar type. Note that the input size was not varied for Feed-Forward Networks because they are not able to process sequences --- they receive the entire 12-letter string as one input.

Our regression analysis found a significant main effect of input size ($z = -51.500, p<0.0001$) as well as significant interaction effects for input size for each grammar type compared to the strictly $k$-local languages (LT: $z = 18.954, p<0.0001$, LTT: $z = 12.998, p<0.0001$, LTTO: $z = 26.281, p<0.0001$, MSO: $z = 100.913, p<0.0001$, CF: $z = 82.311, p<0.0001$, CS: $z = 44.041, p<0.0001$).

Shorter input sizes lead to higher performance for the simpler languages, whereas the Monadic Second-Order (MSO), Context-Free (CF), and Context-Sensitive (CS) languages benefitted from longer input sizes. In the case of the MSO languages, performance was very poor with shorter input sizes. Indeed, since the networks need to identify repeating cycles of sub-strings to correctly distinguish grammatical and ungrammatical strings, it is possible that the length of the string used in our experiments (12) is too short for these neural networks to robustly identify the repeating pattern. For the CF and CS languages, the highest mean percentage was for input sizes 7 and 5 respectively. This is explained by the transitions the networks need to identify grammaticality. In CF languages, strings are constructed either as repeated or mirrored 6-letter substrings, or with $A^nB^n$ grammars. The network need only determine if the transitions between the first and last element of a 7-letter substring are legal transitions. Analogously, in CS languages, strings are constructed according to an $A^nB^nC^n$ grammar, meaning that the network need only determine if the transition between the first and last elements of a 5-letter substring are legal. This is a valid finite-state strategy for accepting strings with fixed lengths.

In general, once again there was a distinction between the simpler languages (SL, LT, LTT, LTTO) and the more complex languages (MSO, CF, CS). Networks were only able to achieve comparable performance on the complex languages when they used larger input windows, which provided them with repeated exposure to the same transitions as they move through the string. Larger input sizes also permitted networks to capitalise on finite-state strategies to evaluate grammaticality in non-regular languages.
\FloatBarrier
\section*{Discussion}
In this study, we performed a principled analysis of the consequences of changing the network topology of an artificial neural network (ANN) for learning grammars with different complexities. We considered network size, architecture, and input size. When we compared cognitive performance between feed-forward and recurruent network types the differences were clear, and had features consistent with a major transition in cognition.

Changing network topology from feed-forward to recurrent (either RNN or GRU) qualitatively changed how the network could process input.  FFNs could process the entire string at once. Recurrent networks could process the entire string at once or as sequences of strings of a certain length. This is a qualitative expansion in the kinds of inputs that can be processed by recurrent compared to feed-forward networks.  

We found a substantial effect of input size on the performance of recurrent architectures, particularly for more complex grammars. Recurrence had the most advantage over FFNs at sequence lengths substantially less than the full string length, and especially for more complex grammars. A recurrent architecture that consumes the whole string at once has no advantage over a simpler FFN. This illustrates a second transitional property: recurrent  and feed-forward networks are equivalent when processing an identical input, but recurrent networks have a larger space of possible input types. 

Our results also show evidence of qualitative transitions in network capacity caused by changes in network topology. In particular, the change from feed-forward to recurrent networks improved performance, particularly on grammars drawn from the higher levels of the hierarchy of formal grammars.   While this result follows theoretically, we emphasise that our results concern not just what networks can do in principle but what they can \emph{learn} in finite time. The change in topology changed what was cognitively feasible for a network of a set size and resources.

Our result is broadly in line with previous work on the learning capacity of artificial neural networks \cite{ackerman2020survey}. However, we note that, contrary to \cite{weiss2018practical}, we  found GRUs to be more powerful than straightforward RNNs. That difference likely stems from a difference in methods. Unlike \cite{weiss2018practical}, we were not concerned with generalisation to longer strings, but rather on the comparative advantage of different architectures with the same input space.  The use of a fixed length allowed FFNs to recognise instances of even quite complex grammars, though at substantially reduced accuracy compared to recurrent architectures.  As far as the specific difference, GRUs are easier to train than traditional RNNs on sequence tasks \cite{ChungEmpirical14}, but it may well be that RNNs under an ideal training regime are ultimately better at generalisation. 

This emphasises that the comparative advantage of transitions will ultimately depend on the details of networks. Given that our networks are idealisations of biological recurrence, the difference in detail is less important than the broad conclusion about the importance of topology.

Transitions in the evolution of cognition \cite{barron2023transitions} and in the evolution of multicellular life \cite{SmithThe-major97} are considered cumulative (each building on the other), but not progressive.  The outcomes of a transition can occupy a very different functional space or niche, but they are not simply better than their predecessors. For example, multicellular life can occupy niches unavailable to unicellular life, but conversely many niches occupied by unicellular life forms are unavailable to multicellular life - especially harsh and extreme environments. In our study recurrent networks (both GRUs and RNNs) offered performance advantages over FFNs of similar size, but the recurrent networks are not just better than FFNs; they are harder to train and the addition of gated recurrent units helped to overcome that difficulty. This is conceptually similar to the ``transition barrier" concept in Major Transitions: the transitional gap is both hard to cross and hard to undo \cite{barron2023transitions}. 

We found no effect (and indeed, a small penalty) for lamination when controlling for other variables such as network size. While a negative result, this emphasises that structural complexity \emph{alone} is not sufficient to improve performance. Lamination, if it is useful, likely requires further mechanisms to coordinate  control flow. This also emphasises that intermediate forms of transitions are not necessarily stable, which is why transitions are both rare and contingently irreversible. It may be that more difficult tasks (such as length generalisation) are necessary to develop a comparative advantage for lamination.

Our negative result from lamination in this study does not mean that the evolution of lamination in animal brains (splitting processing into multiple distinct and degenerate streams) confers no benefit over recurrent or feed-forward cognitive architectures.  The benefits to some changes in network topology may only become apparent in certain tasks and/or under specific resource constraints.  We had previously proposed lamination in animals could confer benefits in the construction of complex task sequences such as song learning or primitive tool construction, and biological networks operate under fundamentally different resource constraints to ANNs \cite{barron2023transitions}. We may need to test other task sets to expose any benefits to lamination in ANNs, as well as operationalise lamination in different ways to better match the biological case.

We also note that the advantage of architectural changes increases for more complex grammars. Simple stimuli can be handled by simple neural networks, and when there is not a comparative advantage there will be no evolutionary gradient. That is likely why, taxonomically, simple neural architectures still vastly outnumber complex ones.  Major transitions are cumulative,  not  {progressive}. There is no intrinsic advantage to having a more complex brain; there is only an advantage to some organisms in some contexts. 
 
That is reflected in our results. Strings from sub-regular grammars capture much of the complexity we would expect to see in natural stimuli. The sub-regular grammars (or, more specifically, the grammars recognizable by counter-free automata) depend on more or less local relationships between features. It is only when an organism need to keep track of more complex spatiotemporal features --- precisely how many times a feature appears, for example --- that the additional complexity becomes important.   The full power of recurrence (and of more complex architectures) is that it enables increasingly sophisticated forms of {memory} for stimuli.  Indeed, even the most powerful computational architectures are equivalent to simple recurrent systems equipped with better memory \cite{GrangerToward20}. But simple, short-lived organisms may have no need to keep track of complex stimuli. To return to a theme we have repeatedly emphasised, the \emph{driver} of a transition must be the contingent  cost-benefit advantage of a more complex architecture.  More sophisticated architectures allow for exploitation of more complex stimuli, but that is an advantage that only accrues downstream of the transition itself. 

\section*{Conclusion}
We show here how some structural changes in artificial networks can confer significant advantages for completing more difficult tasks, such as recognising strings from more complex grammars. This modelling approach lends credence to considering cognitive evolution in terms of structural changes in brains that give rise to changes in computational architectures \cite{barron2023transitions,klein2024comparing}. Major transitions in cognitive evolution do not need to take the form of new cognitive capacities such as imagination or episodic memory. The biggest changes in what is evolutionarily possible may look, at the time, like minor tweaks on existing neural structure. It is an important fact about computational architectures, both natural and artificial, that minimal structural tweaks can lead to major transitions in what it is possible to do. 

\FloatBarrier
\section*{Materials \& Methods}

\subsection*{Artificial Grammars}

Artificial grammars are rule sets, $\{\mathcal{R}\}$ that  specify a set of valid strings, $\{\mathcal{S}\}$ in an artificial language, $\mathcal{L}$. The members of  $\{\mathcal{S}\}$ are strings in a terminal alphabet $\{\mathcal{A}\}$. Rules in $\{\mathcal{R}\}$ are defined over a the set $\{\mathcal{A}\} \cup \{\emptyset, S\}$, where $\emptyset$ is the `empty' element and $S$ is a non-terminal element which can be considered to be a `string under construction.' Lower-case members of the latin alphabet will be used for members of $\{\mathcal{A}\}$ and lower-case Greek letters for strings consisting of both terminal and nonterminal elements.    

\emph{String acceptance} is the problem of determining whether a string $s$ is a member of $\mathcal{L}$. We say that a system computes $\mathcal{L}$ if it can answer, for any string, whether or not it is a member of $\mathcal{L}$.  Fundamental results in computer science \cite{chomsky1959certain, HopcroftAutomata06} show that string acceptance can be used to define equivalence classes of machine architectures, and that these classes fall into a hierarchy, where higher levels can accept strings from a strict superset of the languages of lower levels (see the right hand side of figure \ref{fig:overview}). 

\subsubsection*{Grammar Specification}

We use a six-letter alphabet $\{A\} = \{\text{a}, \text{b}, \text{c}, \text{d}, \text{e}, \text{f}\}$, and produce strings of 12 characters, truncating if necessary.  We do not include $\emptyset$ in $\{A\}$, due to the constraints imposed on these by context-sensitive grammars.  The core component of our grammars are adjacency matrices specifying transitions between members of $\{A\}$. These adjacency matrices specify whether or not there is a valid transition between two characters. A string is initialised with a random letter (or bigram) drawn from $\{A\}$

\subsubsection*{Regular Languages}

Most of the languages we discuss are known as Regular languages. The members of  $\{\mathcal{R}\}$ for a regular language are rules of the kind:

$$S \rightarrow aS;~~~~~ S \rightarrow Sa;~~~~~ S \rightarrow \emptyset$$

The fundamental feature of a rule for regular languages is that the left hand side of the rule contains only one non-terminal and the right hand side contains the combination of one terminal and one non-terminal (in any order).

Any regular languages can be computed by a finite-state automaton. Many regular languages can be recognized by simpler machines, so within the regular language there is a further set of  \emph{subregular} languages.  The subregular language hierarchy has seen extensive use in modelling human and non-human animal communicative systems \cite{rogers2011aural}. 

\paragraph{Strictly k-Local Languages}  $\{\mathcal{S}\}$ consists of strings composed entirely from $k$-grams from some permitted set $T$. String acceptance can be done by a machine which has a  $k$-width sliding window that processes the string left to right.

We set $1 \leq k \leq 3$ and generated adjacency matrices to specify legal transitions. For $k=1$, we randomly select half of the letters in $\{A\}$ and generate strings from random permutations of them. Ungrammatical strings are generated analogously from the complement of the letter set.  For $k=2$, we construct adjacency matrices between each element of the alphabet. Each letter transitions to three other letters with uniform probability. For $k=3$, we construct adjacency matrices between 2-letter bigrams (every permutation) and each letter, again with three transitions per bigram with uniform probability.

\paragraph{Locally $k$-Testable Languages}  A language is locally $k$-testable if it is $k$-local with the added constraint that some subset of $T$ must be present in the string. String acceptance can be done by a machine with a sliding window plus a logic gate to detect if the constraint is satisfied. 

We set $2 \leq k \leq 3$ and generate adjacency matrices to specify legal transitions. We construct the same adjacency matrices as before for these values of $k$, and require that each string is subject to \textit{two} constraints specifying which $k$-length sub-strings must be present in the string at least once to be considered grammatical. The sub-strings in the constraints are randomly sampled from the legal $k$-grams.

\paragraph{Locally $t$-Threshold $k$-Testable Languages} A languaage is locally $t$-threshold $k$-testable if it is $k$-testable and there is some threshold $t>1$ of the maximum number of occurrences any particular substring may occur. Accepting automata need a way to keep count of the number of elements seen. 

We set $2 \leq k \leq 3$ and generate adjacency matrices. We construct the same adjacency matrices as before but additionally  require that each string is subject to \textit{two} constraints specifying which $k$-length sub-strings must be present in the string \textit{only once} to be considered grammatical. The sub-strings in the constraints are randomly sampled from the legal $k$-grams.

\paragraph{Locally $t$-Threshold $k$-Testable With Order Languages} These languages are locally $t$-threshold $k$-Testable Languages with further constraints about which substrings can precede other substrings. String acceptance requires keeping track of not just the number but the order of strings. Computationally, systems which can compute these languages  are known as Counter-Free automata \cite{McNaughtonCounter-Free71, rogers2011aural}.

We set $2 \leq k \leq 3$ and generate adjacency matrices to specify legal transitions. We construct the same adjacency matrices as before for these values of $k$, except that each $(k-1)$-gram can transition to 5 letters rather than 3 so that the languages are more expressive. Each string is again subject to \textit{two} constraints specifying which $k$-length sub-strings must be present in the string \textit{only once} to be considered grammatical. A further constraint is applied that the two sub-strings must be present in a certain (randomly selected) order in the string to be considered grammatical. The sub-strings in the constraints are randomly sampled from the legal $k$-grams.

\paragraph{Second-Order Monadic Languages} These are the most complex regular language. These languages produce strings with repeating substrings which must repeat some number of times \emph{modulo} a number $n$. These languages require the full power of finite-state automata to compute. 

We set $2 \leq k \leq 3$ and generate adjacency matrices to specify legal transitions. We construct adjacency matrices as before, with five transitions between each $(k-1)$-gram for expressivity. Grammars are then generated with Mod-2 and Mod-3 counting. For instance, a grammatical 12-letter string for a Mod-2 grammar could contain 6 2-character repetitions (e.g., $abababababab$), 4 3-character repetitions (e.g., $abcabcabcabc$), or 2 6-character repetitions (e.g., $abcdefabcdef$). In contrast, ungrammatical strings would contain 3 4-character repetitions (e.g., $abcdabcdabcd$). For Mod-3, grammatical strings could contain 6 2-character repetitions or 3 4-character repetitions, while ungrammatical strings contain 4 3-character repetitions or 2 6-character repetitions.

\paragraph{Context-free Languages}

The rule set $\{\mathcal{R}\}$, for a context-free language contains rules of the kind:

$$S \rightarrow \omega$$

The fundamental feature of a context-free language is that the left-hand-side of the rule contains only a single non-terminal and that the right-hand-side contains a string of terminals and non-terminals. 

We generated adjacency matrices for specifying legal transitions between characters in $\{\mathcal{A}\}$. We produced three types of context-free grammars. \textit{Repeated} grammars sampled half the string from the adjacency matrix and then repeated it for grammatical strings, it mirrored it for ungrammatical strings. \textit{Mirrored} grammars sampled half the string and then mirrored it for grammatical strings and repeated it for ungrammatical strings. From the perspective of a classification task, these problems are equivalent as the decision manifold is identical, \textit{ceteris paribus}. We also produced $A^nB^n$ grammars. Considering a 1-indexed string of length 12, a letter is sampled at random and placed in position 1. A valid transition from it is then sampled from the adjacency matrix and placed at position 7. We repeat this process for positions 2 and 8, 3 and 9, and so forth. Ungrammatical strings are generated from the complement of the adjacency matrix, so the long-distance dependencies are different even if local transitions are the same.

\paragraph{Context-Sensitive Languages}

The rule set $\{\mathcal{R}\}$, for a context-sensitive language contains rules of the kind:

$$\chi S\psi \rightarrow \omega$$

The fundamental identifier of a context-sensitive language is that both sides can contain a mixture of terminals and non-terminals.

We produce only one type of context-sensitive grammar, $A^nB^nC^n$ grammars. The processing for sampling strings is the same as $A^nB^n$ languages, except that the process is repeated for a further iteration. Considering a 1-indexed string of length 12, a letter is sampled at random and placed in position 1. A valid transition from it is then sampled from the adjacency matrix and placed at position 5. A valid transition from that letter is then sampled from the adjacency matrix and placed at position 9. We repeat this process for positions (2, 6, 10), (3, 7, 11), and (4, 8, 12). Ungrammatical strings have the transitions sampled from the complement of the adjacency matrix, again so that the long-distance dependencies are different even if local transitions are the same.

\subsection*{Artificial Neural Networks}

We systematically create $800$ artificial neural networks (ANNs) with different properties, corresponding to three of the five proposed transitions \cite{barron2023transitions}. We create fully-connected feed-forward ANNs, laminated feed-forward ANNs, fully-connected recurrent ANNs, and laminated recurrent ANNs. For recurrent networks, we also vary the input size of the models, meaning the number of characters they see per hidden state update. 

\subsubsection*{Feed-Forward Networks}

These networks take an encoded representation of the string and return a single probability associated with a class (grammatical, ungrammatical). The string is represented as a concatenation of twelve one-hot encoded vectors of length six representing each character, resulting in a $72\times 1$ vector input.

\subsubsection*{Recurrent Neural Networks}

These networks take sequences of one-hot encoded representations of the sub-strings in a string. For models seeing only one character at a time, they receive a $6\times 1$ vector, for models seeing two characters at a time, they receive a concatenation of two one hot-encoded vectors, and so forth up until a model that sees 12 characters at a time (i.e., the whole string). The model iterates through the string as a sequence, moving one character forward at a time. Thus, the model proceeds through the string like a sliding window. We update the weights of the recurrent component of the network as it receives each input, updating its hidden states. Once the whole sequence has been given to the network, we then use a single dense layer at the end to predict the probability of the class. We implement both standard recurrent layers and gated recurrent layers. Gated recurrent networks are identical to vanilla recurrent networks, except that they include two gates which modulate how much previous states should affect the next state of the network, helping to minimize the problems of vanishing gradients during fitting.

\subsubsection*{Lamination}

To approximate lamination, we introduce partitions into the neural networks, whereby the hidden layers of each network are partitioned and not fully connected, leading to parallelised information flow down independent channels in the network, before being recombined in the final layer.

\subsection*{Training Details}

We varied the number of connections in each network between 32 and 512 inclusive in increments of 32, and varied the number of hidden layers in each network between 1 and 3 inclusive. We also introduce 1 or 2 laminations, where 1 lamination is the fully connected network.

We use Rectified Linear Units as activation functions until the final layer, which has a sigmoid activation function. We train for a maximum of 100 epochs using batch sizes of 100. We train on 70\% of the strings for a given grammar and evaluate on the remaining 30\% after every batch of every epoch. We use the Brier Score and the percentage of correctly classified strings to evaluate each network on the test set. The Brier Score is defined as  $$\frac 1 N \sum^N_{i = 1}[(\hat{y}_i - \tilde{y}_i)^2]$$ where $\tilde{y}_i$ is the target class for string $i$ and $\hat{y}_i$ is the predicted probability that the string is ungrammatical \cite{brier1950verification}. Percentage correct is simply  $$\frac{\sum^N_{i=1}\hat{y}_i \geq 0.5 \equiv \tilde{y}_i}{N}\cdot100$$ If this percentage reaches 100\%, training is stopped. 
Networks were trained with backpropagation, with the binary cross entropy loss and the momentum optimizer ($\eta = 0.01,~\rho = 0.95$).

\subsection*{Statistical Analysis}

To analyse this data, we use a fixed effects Beta regression \cite{ferrari2004beta} and conduct parallel analyses both on the proportion of test strings correctly classified and on the Brier score on the test set. We found no interpretative difference between using proportion and Brier Score in the regression models, and so we have proceeded with reporting proportion outcomes (transformed to percentages) in the main text for ease of interpretation. We used Smithson-Verkuilen shrinkage to handle 0 or 1 response values \cite{smithson2006better}. We conducted extensive model selection to arrive at our final model formulation for both the main effects and dispersion submodels:

\begin{equation*}
    \text{R} \sim \text{A}*\text{G} +
    \text{L}*\text{G} +
    \text{I}*\text{G} +
    \text{A}*\text{L} +
    \text{N} + 
    \text{D}
\end{equation*}

where $*$ denotes an interaction and composite main effects for each operand. $R$ is the response variable, $A$ is the architecture factor (FFN, RNN, GRU), $G$ is the grammar factor (SL, LT, LTT, LTTO, MSO, CF, CS), $I$ is the numerical input size, $L$ is the laminations factor (Dense, Laminated), $N$ is the number of neurons, and $D$ is the depth of the network (number of layers). This model achieved the lowest Akaike Information Criterion, Bayesian Information Criterion, and 5-fold cross-validated root mean-squared error of all candidate models. We used log-log links for the main effects model and log links for the dispersion model \cite{ferrari2004beta}. We used $\alpha = 0.05$ as our significance threshold, and report estimated marginal means and the significance of pairwise comparisons from the fitted models in the main text. All data, code, and analysis can be found in the \href{https://github.com/MatisPatel/AGLProj}{supplementary material}.

\section*{Acknowledgments}
This work was supported by the Templeton World Charity Foundation (TWCF-2020-20539 to AB, MH, and CK) and by the Australian Research Council (DP240100400 to CK and AB).

\section*{Contributions and Declarations}
MP conceived the project; KV conducted all experiments and analyses; all authors contributed to the original and revised drafts. The authors declare no competing interests relevant to this work. Correspondence should be addressed to KV: \texttt{kv301@srcf.net}.
\bibliographystyle{unsrt}  
\bibliography{references}

\end{document}